\documentclass[journal,twoside,web]{ieeecolor}

\usepackage{etoolbox}
\makeatletter
\@ifundefined{color@begingroup}%
{\let\color@begingroup\relax
\let\color@endgroup\relax}{}%
\def\fix@ieeecolor@hbox#1{%
\hbox{\color@begingroup#1\color@endgroup}}
\patchcmd\@makecaption{\hbox}{\fix@ieeecolor@hbox}{}{\FAILED}
\patchcmd\@makecaption{\hbox}{\fix@ieeecolor@hbox}{}{\FAILED}

\usepackage{tmi}
\usepackage{cite}
\usepackage{amsmath,amssymb,amsfonts}
\usepackage{algorithmic}
\usepackage{graphicx}
\usepackage{textcomp}
\usepackage{mathrsfs}
\usepackage{url}
\usepackage{multirow}
\usepackage{makecell}
\usepackage{booktabs}
\usepackage{hyperref}
\usepackage{lineno}

\begin{document}
\title{Towards Better Cephalometric Landmark Detection with Diffusion Data Generation}
\author{Dongqian Guo, Wencheng Han, Pang Lyu, Yuxi Zhou, and Jianbing Shen, \IEEEmembership{Fellow, IEEE} \\
~\thanks{
Dongqian Guo and Wencheng Han have equal contributions.
Corresponding author: \textit{Jianbing Shen}.
This work was supported in part by the Jiangyin Hi-tech Industrial Development Zone under the Taihu Innovation Scheme (EF2025-00003-SKL-IOTSC), the University of Macau SRG2022-00023-IOTSC grant, and the Shanghai Science and Technology Development Funds (24YF2704700).
}
~\thanks{Dongqian Guo, Wencheng Han, and Jianbing Shen are with the State Key Laboratory of Internet of Things for Smart City, Department of Computer and Information Science, University of Macau, Macau.}
~\thanks{
Pang Lyu is with Department of Orthopaedic Surgery, Zhongshan Hospital, Fudan University, Shanghai, 200032 China.}
~\thanks{Yuxi Zhou is with the Department of Periodontology, Justus-Liebig-University of Giessen, Germany, the Department of Periodontics, Stomatology Hospital of Guangzhou Medical University, China.}
}

\maketitle
\begin{abstract}
Cephalometric landmark detection is essential for orthodontic diagnostics and treatment planning. {Nevertheless, the scarcity of samples in data collection and the extensive effort required for manual annotation have significantly impeded the availability of diverse datasets.} This limitation has restricted the effectiveness of deep learning-based detection methods, particularly those based on large-scale vision models.
To address these challenges, we have developed an innovative data generation method capable of producing diverse cephalometric X-ray images along with corresponding annotations without human intervention. To achieve this, our approach initiates by constructing new cephalometric landmark annotations using anatomical priors. Then, we employ a diffusion-based generator to create realistic X-ray images that correspond closely with these annotations.
To achieve precise control in producing samples with different attributes, we introduce a novel prompt cephalometric X-ray image dataset. This dataset includes real cephalometric X-ray images and detailed medical text prompts describing the images. By leveraging these detailed prompts, our method improves the generation process to control different styles and attributes. 
Facilitated by the large, diverse generated data, we introduce large-scale vision detection models into the cephalometric landmark detection task to improve accuracy.  Experimental results demonstrate that training with the generated data substantially enhances the performance. 
Compared to methods without using the generated data, our approach improves the Success Detection Rate (SDR) by 6.5\%, attaining a notable 82.2\%. \textit{All code and data are available at: \url{https://um-lab.github.io/cepha-generation/}}
\end{abstract}


\section{Introduction}
\label{sec:introduction}

Detecting landmarks on X-ray images of the craniofacial area is pivotal for orthodontic treatment~\cite{kielczykowski2023application, rauniyar2023artificial}. This task requires precise identification of anatomical landmarks on cephalometric X-rays to evaluate facial structure and growth, plan orthodontic treatment, and monitor post-treatment changes. Despite its critical importance, this field encounters several challenges. A noteworthy issue is the limited scale of available training datasets, typically consisting of only a few hundred samples. This scarcity significantly hampers the generalization capabilities of detection models. Consequently, there is a pressing need to develop strategies that can enhance the performance and reliability of landmark detection in cephalometric images.

\begin{figure}[t]
\centering
    \includegraphics[scale=0.6]{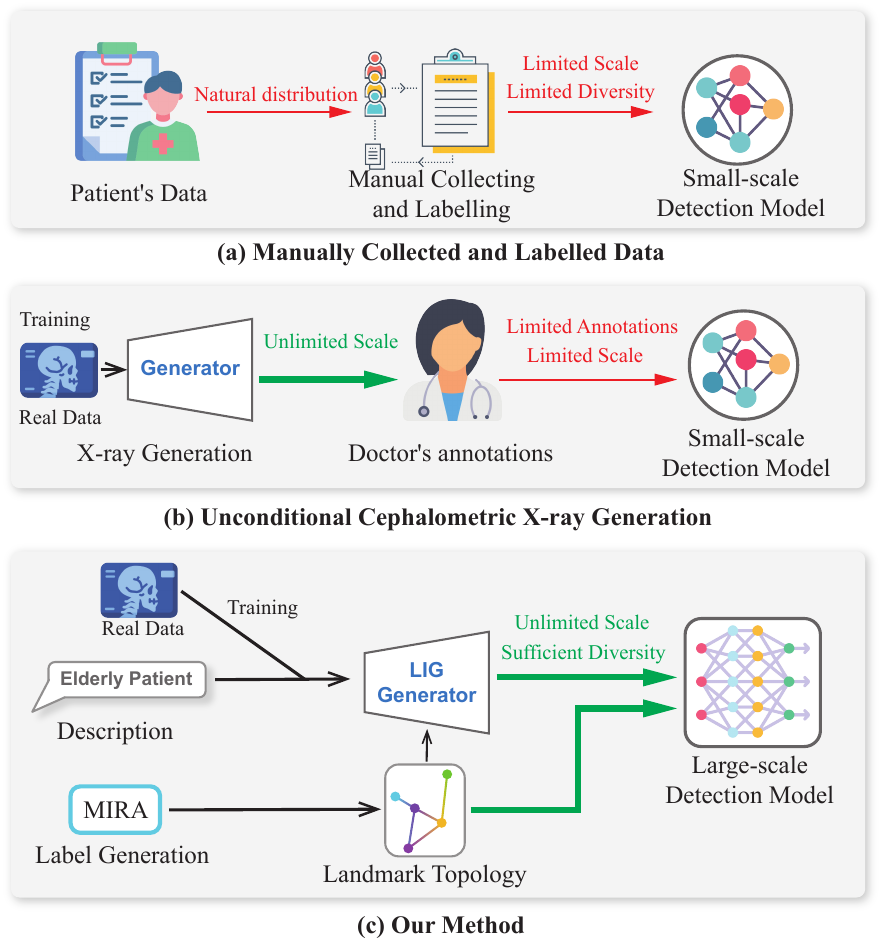}
    \caption{\textbf{The Comparison of Traditional and Our Method.} The lack of samples and complex labeling process limit the data scale and diversity. To address this limitation, we propose a conditional data generation method that can simultaneously generate X-ray images and corresponding annotations. This approach enables incorporating large-scale detection models into this area without introducing overfitting.}
    \label{fig:landmark_motivation}
\end{figure}

As shown in Figure~\ref{fig:landmark_motivation}(a), cephalometric X-ray images in this field come from the manual collection of patient radiographs. The natural distribution of patient data over time limits diversity and makes data collection challenging. The problem is further exacerbated in corner-case scenarios, where patient samples are rare, making collecting sufficient data even more difficult. Additionally, annotating these images is another challenge. Identifying cephalometric landmarks requires extensive medical expertise, making it difficult for non-experts to annotate accurately, which creates high workloads for experienced medical professionals and limits the ability to scale. To our knowledge, there are only a few open-source datasets for cephalometric landmark detection, each containing at most hundreds of images~\cite{wang2016benchmark, wang2015evaluation}.

Given these data constraints, the choice of models for landmark detection is also limited. Large-scale vision detection models are generally ineffective with small-scale training data, as these parameter-rich methods tend to overfit by memorizing the distribution of the small dataset. Therefore, small-scale detection models such as HRNet~\cite{wang2020deep} are preferred.

To address the challenge of data scarcity, some methods propose generating additional X-ray images. For instance, Madani ~\textit{et. al.}~\cite{madani2018chest} utilize Generative Adversarial Networks (GANs) to create new chest X-ray samples, improving the classification of normal and abnormal samples. 
Similarly, Karbhari \textit{et. al.}~\cite{karbhari2021generation} employ Auxiliary Classifier GANs (ACGANs) to generate chest X-rays for diagnosing COVID-19, significantly advancing AI diagnostic capabilities. 
While these methods have shown success in classification tasks, they do not resolve the data scarcity for cephalometric landmark detection. As shown in Figure~\ref{fig:landmark_motivation}(b), while unconditional generation methods can produce more images, they cannot provide precise landmark annotations. This limitation restricts the use of generated training data in this field.

{To totally solve these problems, we introduce the Anatomy-Informed Cephalometric X-ray Generation (AICG) pipeline, depicted in Figure~\ref{fig:landmark_motivation}(c). Our method generates detailed cephalometric X-ray samples with their corresponding landmarks, eliminating manual annotation and allowing almost unlimited sample generation. To achieve this, we first create new landmark annotations using medical prior knowledge and then use these annotations as control signals to generate the X-ray images. 
To facilitate the generation process given the landmark conditions, we integrate anatomical knowledge into ControlNet and develop a topology module that maps relationships between different landmarks. 
Additionally, we propose a new cephalometric X-ray dataset with detailed medical descriptions of each patient, linking cephalometric structures to text descriptions. These annotations help our pipeline generate diverse samples with varying attributes and features, which is essential for increasing rare or unusual cases.}

Using the large variety of generated training data, we implement advanced large vision detection models for this task. These models significantly improve the accuracy of landmark detection, promoting its application in real medical scenarios. In summary, the contributions of this paper can be encapsulated in four points:
\begin{itemize}
   \item \textbf{Anatomy-Informed Cephalometric X-ray Generation Pipeline:} We propose a novel pipeline that generates both cephalometric X-ray images and their corresponding landmark labels. To the best of our knowledge, this is the first work to solve both the data acquisition and labeling problems in the cephalometric landmark detection area.

    \item \textbf{Cephalometric Anatomy-Informed Topology Module:} We propose a new topology module that reflects cephalometric anatomy based on medical knowledge. This module enhances understanding of landmark relationships, helping the generation model accurately depict cephalometric structures that correspond to landmark labels.
 
    \item \textbf{Prompt-Cephalometric X-ray (Prompt-CX) Dataset:} {We introduce a new cephalometric X-ray dataset that pairs real X-ray images with detailed medical descriptions of patients. This is the first dataset to combine cephalometric X-ray image generation with medical text descriptions, helping to understand the relationship between X-ray images and language descriptions of patients.}

    \item \textbf{Large-scale Detection Models of Cephalometric Landmark Detection:} {
    Leveraging large generated data, we incorporate large-scale detection models into this area without concerns about overfitting to small-scale training data. Compared to previous methods, our approach significantly improves the Success Detection Rate (SDR) by 6.5\%, reaching 82.2\%. }
\end{itemize}

\section{Related Works}
\label{sec:related_works}
\subsection{Generative Models in the Medical Fields}
The development of generative models enables us to address the scarcity of medical imaging data through synthetic data augmentation~\cite{04, huang2024diverse, garcea2023data, pezoulas2024synthetic}. In the early years, GANs~\cite{goodfellow2014generative} pioneered the generation of medical images~\cite{03, kwon2019generation}.
However, an empirical study proposed by Skandarani \textit{et al.}~\cite{skandarania2021gans} concludes that the GANs are challenging to achieve the full richness of medical datasets although they are capable of generating realistic-looking medical images when they test GANs on cardiac cine-MRI, liver CT, and RGB retina images.

The proposal of diffusion models~\cite{rombach2022high} has revolutionized the field of image generation. 
To address the limitations of GANs, many researchers now adopt diffusion models to synthesize medical training data for improving model performance of specific tasks\cite{chen2021synthetic, pengfeiguo2024addressing}, such as detection of chest X-ray lesions~\cite{weber2023cascaded, shentu2024cxr}, simulating the disease progression~\cite{liang2023pie} and some cross-modality images translation and generation tasks~\cite{05, zhan2024medm2g}.
Recently, the ControlNet~\cite{zhang2023adding} allows us to not only use text but also use various conditions to finely control the generation process, such as using soft edges and tiled images as conditions to generate the images during the Laparoscopic Cholecystectomy~\cite{kaleta2023lc}. Also, some frameworks such as MONAI~\cite{pinaya2023generative} are developed to assist in generating medical image data.

\label{sec:cephalometric_X-Ray_image_generation}
\begin{figure*}[t]
\centering
    \includegraphics[scale=0.98]{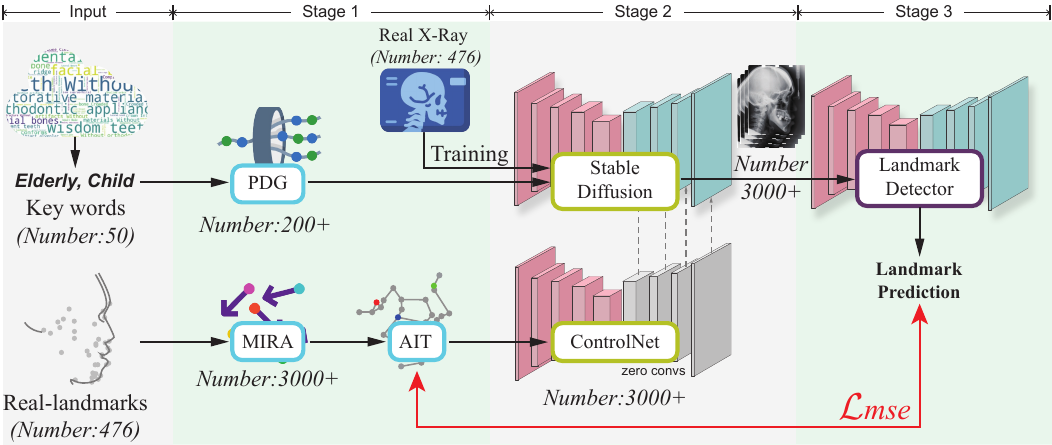}
    \caption{\textbf{Overview of the Anatomy-Informed Cephalometric X-ray Generation (AICG) Framework.} 
    Three primary stages include Condition Generation (highlighted in blue), Image Generation (marked in yellow), and Landmark Detection (stroked in purple), delineating the pipeline from condition preparation through image synthesis to landmark detection.}
    \label{fig:landmark_structure}
\end{figure*}

\subsection{Cephalometric Landmark Detection}
Traditional learning strategies require researchers to manually integrate systematic knowledge, aiming to create reliable guidance for the models~\cite{levy1986knowledge}. However, the growing complexity of images and underlying mechanisms throw significant challenges on such methods. Many efforts have been made to improve the detection accuracy and better identify the features from cephalometric X-rays.
For example, some edge and pattern detection-based methods are proposed~\cite{mosleh2008image, pouyan2010cephalometric}, but these models are not widely adaptable because their performance usually fluctuates significantly in different application scenarios. Later, machine learning-based methods such as random forest are raised and gradually surpass traditional methods in this field~\cite{ibragimov2014automatic},~\cite{tim2015fully, lindner2016fully}. Recently, the deep learning based techniques~\cite{wu2023revisiting, lee2019deep},~\cite{chen2019cephalometric} become the mainstream. For instance, Cephalformer~\cite{jiang2022cephalformer} leverages the relationships between visual concepts and landmarks to offer valuable guidance for precise detection of cephalometric landmarks in both 2D and 3D. Some few-shot~\cite{quan2022images} and one-shot~\cite{zhu2023uod} methods are proposed to relieve the lack of data. However, they focus on labeling fewer images without compromising training results and maximizing the extraction of key features from limited annotations.
A large amount of data is crucial to ensure the performance of models in detecting cephalometric landmarks. When data scales are under expectation, complementary annotation are often applied, though it indeed brings heavy workloads for improving the accuracy of detection models. Hence, generating high-quality images with rich diversity will help solve these challenges.

\section{Cephalometric X-ray Diffusion Generation and Landmark Detection}
\subsection{Overview}
\label{sec:cephalometric_X-Ray_image_generation:a}
As shown in Figure~\ref{fig:landmark_structure}, our Anatomy-Informed Cephalometric X-ray Generation (AICG) pipeline comprises three primary stages.
The first stage is the Condition Generation Stage, and its purpose is to set up the conditions needed for image generation, including landmark and text descriptions.
The landmark conditions generate labels based on anatomical prior knowledge, guiding the generator to create X-ray images that match these labels. These labels will serve as ground truth for landmark detection models. Text descriptions help the image generator produce X-rays with varying attributes, such as those from different scanners. While images may share the same landmark positions, they can show notable differences based on their descriptions. The key modules in this stage include the Anatomy-Informed Random Augmentation (MIRA) Module, the Anatomy-Informed Topology (AIT) Module, and the Prompt Description Generator (PDG).

The MIRA Module takes a small set of real landmark data $L$ (476 images from our training set) as input. 
Then, a series of Global Augmentation and Anatomy-Informed Augmentation techniques are applied to randomly generate a diverse and anatomically consistent set of landmark information:
\begin{equation}
    L' = \text{MIRA}(L, n_l),
\end{equation}
where $n_l$ is the augmented landmark number. In our experiments, this process totally generates more than 3000 different landmark labels. The specific design of the MIRA module is further elaborated in Section~\ref{sec:cephalometric_X-Ray_image_generation:b}.

The AIT module transforms the position information of landmarks by MIRA into Anatomy-Informed Topology labels:
\begin{equation}
    T = \text{AIT}(L').
\end{equation}

These labels include precise locations of landmarks and the anatomical relationships between different landmarks, aiding subsequent generation modules in understanding the features encoded by the landmarks. 
The AIT module will be detailed in Section~\ref{sec:cephalometric_X-Ray_image_generation:c}. 
The Prompt Description Generator (PDG) takes feature keywords $K$ annotated by experienced doctors (50 different keywords used in our experiments) and generates over 200 different prompt descriptions through conditional permutation to control image generation:
\begin{equation}
    D = \text{PDG}(K).
\end{equation}

The specific design of the PDG module will be introduced in Section~\ref{sec:cephalometric_X-Ray_image_generation:d}.
The second stage is the Image Generation Stage as shown in Figure~\ref{fig:landmark_structure}. 
The generator design will be discussed in detail in Section~\ref{sec:cephalometric_X-Ray_image_generation:e}.
This stage inputs conditions generated from the previous stage and random Gaussian noise, outputting corresponding X-ray Images:
\begin{equation}
X = \text{Generator}(T, D, \mathcal{N}).
\end{equation}

The final stage is the Landmark Detection Stage. Previous landmark detectors are generally based on networks with relatively small parameter sizes due to the limited scale of data in this field. However, with the generation of thousands of accurately labeled samples in the proposed method, we can now use more extensive, more powerful large-scale vision detection models (LVM) to handle this challenging task:
\begin{equation}
    \theta^* = \arg\min_\theta \sum_{(x_i, y_i) \in X} L(\text{LVM}_\theta(x_i), y_i) .
\end{equation}
These new detectors will be described in Section~\ref{sec:cephalometric_landmark_detection_method}.

\subsection{Cephalometric Anatomy-Informed Random Augmentation (MIRA)}
\label{sec:cephalometric_X-Ray_image_generation:b}
The traditional process of collecting and annotating cephalometric X-ray images requires significant time and effort. Although image generation techniques have been applied in other medical fields to create more diverse samples, cephalometric landmark detection demands more precise annotation of 38 distinct landmarks, which require specialized medical knowledge to identify accurately.
To address these challenges, we propose a data generation method that generates both cephalometric X-ray images and their landmark annotations. Specifically, our method first generates different landmark labels, which are then used as control signals to create cephalometric X-ray images that align with these labels. This approach avoids the labor-intensive manual annotation while guaranteeing that the generated images preserve anatomically accurate structures.

\begin{figure}[t]
\centering
    \includegraphics[scale=0.48]{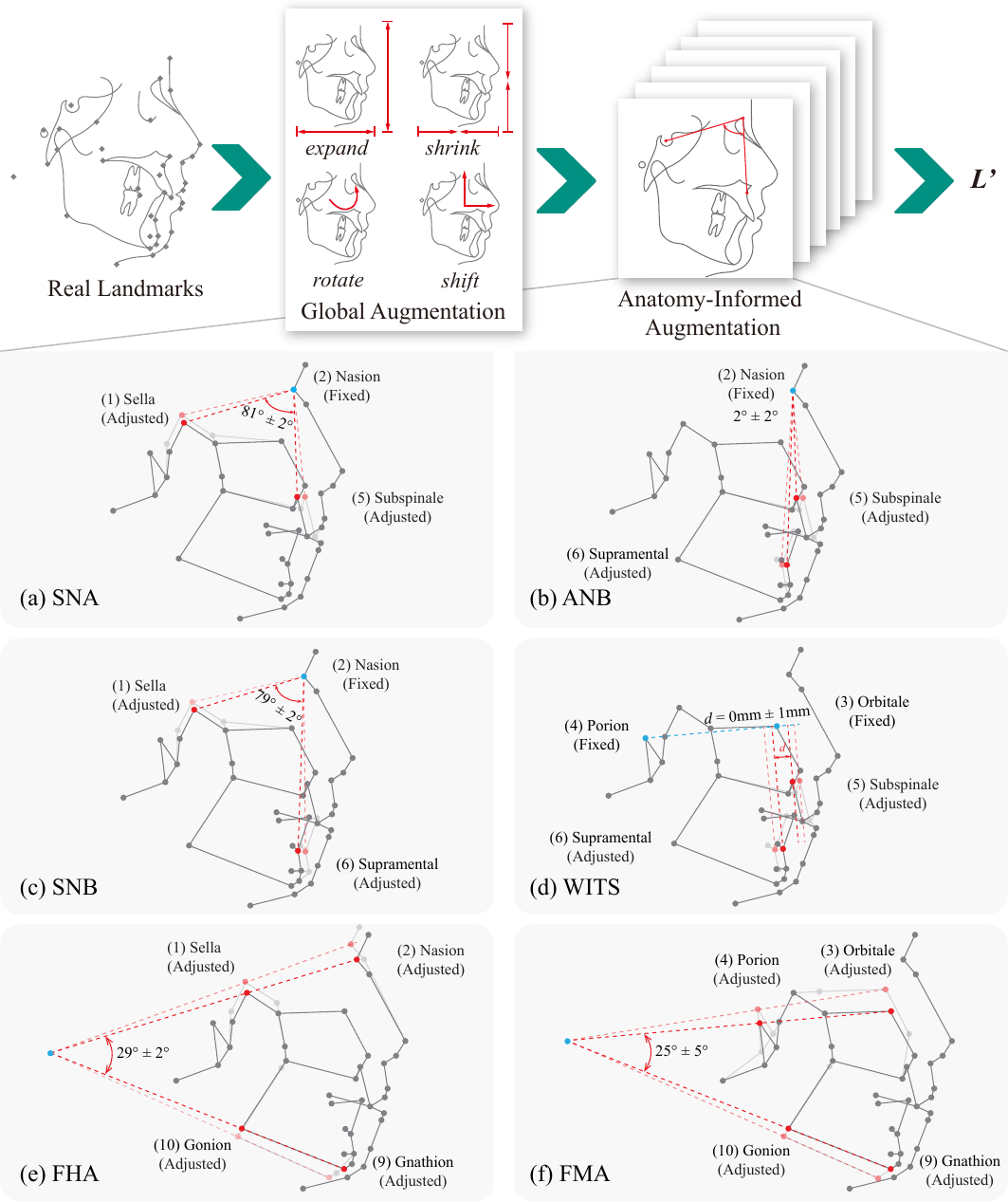}
    \caption{\textbf{The Proposed MIRA Module.} This figure depicts the transformation of real landmark labels through Global Augmentation and the rules of Anatomy-Informed Augmentation to generate a diverse and anatomically accurate position of cephalometric landmarks.}
    \label{fig:Landmark_MIRA}
\end{figure}

To generate a large and diverse array of landmark labels that comply with anatomical priors, we propose the Anatomy-Informed Random Augmentation (MIRA) Module. As shown in Figure~\ref{fig:Landmark_MIRA}, this module takes a small set of real landmark labels as input and then applies a series of global and anatomy-informed enhancements to generate a large set of new landmark labels.
Specifically, this process randomly expands the set of landmark groups $L=\{l_1, l_2,...,l_{38}\}$ into several new groups.
In this process, global augmentations are first applied, affecting all landmarks uniformly without altering their relative positions.
This includes random expansions and shrinkages in the X and Y directions to simulate the distribution of landmarks across different body sizes, as well as random rotations and overall shifts in the XY direction, mimicking varying patient positions and scanner placements during imaging. All global augmentations can be integrated into a single affine transformation matrix to accomplish:
\begin{equation}
    l^{'}_i = A\cdot l_i.
\end{equation}
\begin{align}
&A=
\begin{bmatrix}
x_i' \\
y_i' \\
1
\end{bmatrix}
&=
\begin{bmatrix}
s_x \cdot \cos(\theta) & -s_y \cdot \sin(\theta) & t_x \\
s_x \cdot \sin(\theta) & s_y \cdot \cos(\theta) & t_y \\
0 & 0 & 1
\end{bmatrix}
\cdot
\begin{bmatrix}
x_i \\
y_i \\
1
\end{bmatrix} 
\end{align}
where A is the affine transformation matrix.
$s_x$ and $s_y$ are the scaling factors for expansion or shrinkage in the X and Y directions. $\theta$ is the rotation angle. $t_x$ and $t_y$ are translation (shifts) in the X and Y directions.

As shown in Figure~\ref{fig:Landmark_MIRA}, we also develop six anatomy-informed augmentations grounded in anatomical theory~\cite{tweed1946frankfort, steiner1953cephalometrics, wang2016benchmark, paddenberg2021floating}. Each augmentation is based on well-established relationships between key anatomical landmarks, such as angles and distances. When generating new landmarks, we randomly select one or more of these augmentations and apply them after the Global Augmentation. For example, in the (a) SNA augmentation, we randomly adjust the angle between the L-1 Sella and L-5 Subspinale, using the L-2 Nasion axis. The angle is constrained within a range of 79-83 degrees. In addition to moving the Sella and Subspinale, we also adjust the neighboring landmarks that are directly associated with them, resulting in a new relative positioning of the landmarks.
This augmentation method aligns with the anatomical relationships summarized in the existing theory, ensuring that the augmented landmarks remain consistent with real human anatomical distributions.

\subsection{Cephalometric Anatomy-Informed Topology (AIT)}
\label{sec:cephalometric_X-Ray_image_generation:c}

\begin{figure*}[t]
\centering
    \includegraphics[scale=0.41]{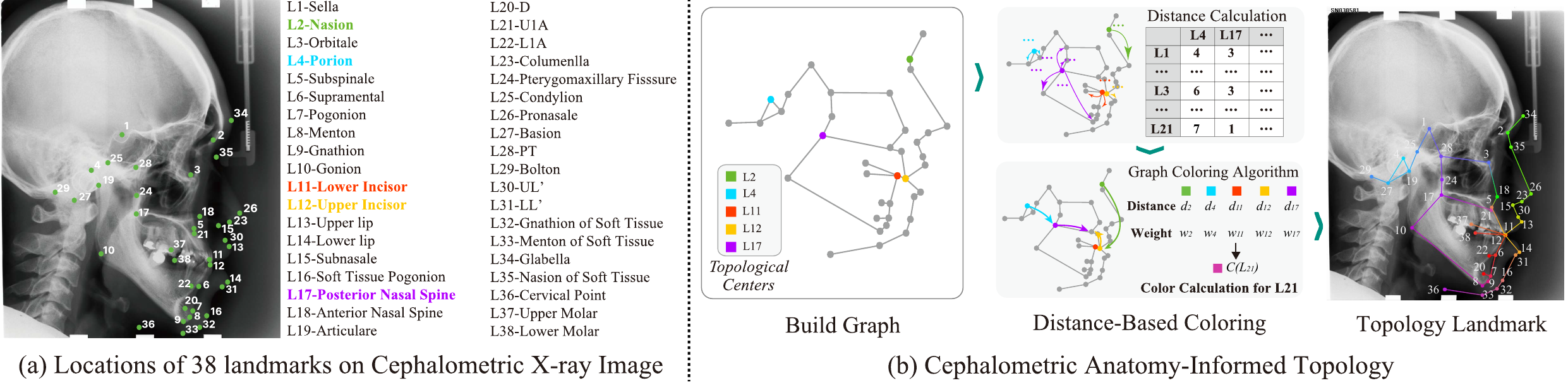}
    \caption{\textbf{The Anatomy-Informed Topology (AIT) Module:} (a) Shows 38 cephalometric landmarks with their positions and names. (b) Describes the AIT module's process, highlighting the construction of a graph with critical landmarks and employing Distance-Based Coloring for intuitive anatomical relationship representation.}
    \label{fig:landmark_topology}
\end{figure*}

To facilitate accurate alignment with input annotations, we integrate these landmarks as conditions into the generative network. The next challenge is how to efficiently input these landmarks into the image generator.

The most straightforward method is to plot all the landmarks on a single-channel image and use this image as input. However, this method may lead to misinterpretations by the model due to the variability in the relative positions of landmarks across different individuals. For example, landmarks L-11 and L-12, which represent the lower and upper incisor, might have completely different relative positions in different individuals due to varying bite alignments. Therefore, a single-channel landmark image can be ambiguous in many situations, making it an unreliable basis for generation. An alternative approach could be a multi-channel image, where each channel represents a different landmark. However, this representation can make it difficult to understand the relationships between various landmarks. Furthermore, since each channel contains relevant information in less than 1\% of its area, it poses a challenge to the generative model to learn the correlation between landmarks and the final image effectively.

To address these issues, we introduce a new representation method: the Cephalometric Anatomy-Informed Topology (AIT) module in Figure~\ref{fig:landmark_topology} (b).
{First, we construct a graph $G$ based on anatomical principles, converting the 38 discrete landmarks into a coherent graph. In this graph, we set five landmarks \(\mathcal{L} = \{L_2, L_4, L_{11}, L_{12}, L_{17}\}\) as topological centers because of their key roles and assign them unique colors \(\mathbf{C(\mathcal{L})}\). We then use a distance-based coloring algorithm to determine the colors for the remaining nodes based on their distances to these critical nodes. The graph distance between each remaining node \(v\) and the critical landmarks is \(d(v, L)\), and the weight \(w_L(v)\) is the reciprocal of \(d(v, L)\). 
Then, we normalize the weights}:
\begin{equation}
    \hat{w}_L(v) = \frac{w_L(v)}{\sum_{L'\in\mathcal{L}}w_{L'}(v)}
\end{equation}

Finally, we mix the colors of the critical landmarks to obtain the final color \(\mathbf{C}(v)\) of the node \(v\):
\begin{equation}
    \mathbf{C}(v) = \sum_{L\in\mathcal{L}}\hat{w}_L(v)\mathbf{C}(L)
\end{equation}

We then assign gradient-colored edges to connect the nodes. Let the color of the two nodes of the edge be \(\mathbf{c_1}, \mathbf{c_2} \in \mathbb{R}^3\). We generate the gradient colors based on the following rules:
\begin{equation}
    \mathbf{c}(t) = \mathbf{c}_1 + \frac{t}{D - 1}(\mathbf{c}_2 - \mathbf{c}_1), t = 0,1,...,D - 1.
\end{equation}
where \(D\) is the pixel distance between two nodes. The resulting gradient color sequence is: \(\mathbf{c}(0), \mathbf{c}(1), ..., \mathbf{c}(D - 1)\), and \(\mathbf{c}(0) = \mathbf{c}_1, \mathbf{c}(D - 1) = \mathbf{c}_2\).

As shown in the figure, the connections and colors of the nodes and edges intuitively depict the relationships between different nodes. This representation simplifies the model's understanding of the anatomical information in the landmarks.

\subsection{Cephalometric Prompt Description Generator (PDG)}
\label{sec:cephalometric_X-Ray_image_generation:d}
Using landmarks as control conditions allows us to manipulate the X-ray image generator to align with the given landmarks. However, certain aspects of the images, unrelated to landmark positioning, remain uncontrollable. These include factors such as scanner types and patient attributes such as features of the oral region. We refer to these uncontrollable aspects as side conditions. Although they do not directly influence the position of the landmarks, they significantly impact the robustness of the detection models. To ensure that our detection models fully understand different side conditions, we use text prompts integrated with landmark topology information to control image generation.

\begin{figure}[t]
\centering
    \includegraphics[scale=0.98]{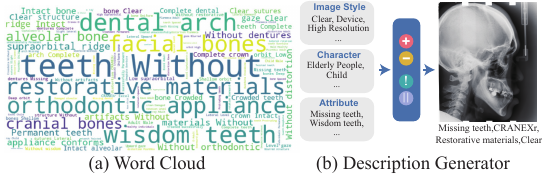}
    \caption{\textbf{The Cephalometric Prompt Description Generator.} (a) A word cloud depicts the commonality of keywords annotated by medical experts. (b) The Prompt Description Generator module's schematic creates diverse image descriptions across three categories.}
    \label{fig:landmark_prompt}
\end{figure}

We engage five experienced medical experts for this project, who are assigned the task of annotating each X-ray image using different keywords. Figure~\ref{fig:landmark_prompt} (a) displays a word cloud of the annotated keywords. The size of each word represents its frequency of occurrence in all samples. 
Some of these features are infrequent, but the strong generalization and language understanding capabilities of large-scale visual-language pre-trained models like CLIP~\cite{radford2021learning} allow our generative model to easily generalize these low-frequency features across a broader range of samples.

To enrich the diversity of prompts in image generation, we develop a Prompt Description Generator module. 
As shown in Figure~\ref{fig:landmark_prompt} (b), the prompt descriptions are categorized into three groups. The Image Style group covers overall image characteristics, such as clarity and scanner type. The Character group describes patient characteristics, such as age (\textit{e.g., child or elderly}). The Attribute group focuses on specific patient features, such as \textit{Missing Teeth} or \textit{Wisdom Teeth}. To ensure logical consistency in the generated descriptions, we design several rules between these groups. For instance, \textit{deciduous teeth} should not be combined with \textit{adult}, nor should \textit{wisdom teeth} be paired with \textit{child}, and so on. During the prompt generation process, we randomly select one description from both the Image Style and Character groups and choose several attributes from the Attribute group to form the final prompt.

This text prompt control mechanism enhances the data generation process by enabling the model to produce a diverse range of anatomically precise and contextually rich images. These images capture a broader spectrum of clinical scenarios, thereby improving the robustness and generalizability of our cephalometric detection models.

\subsection{Cephalometric X-ray Images Generation Network}
\label{sec:cephalometric_X-Ray_image_generation:e}

\begin{figure}[t]
    \centering
    \includegraphics[scale=0.75]{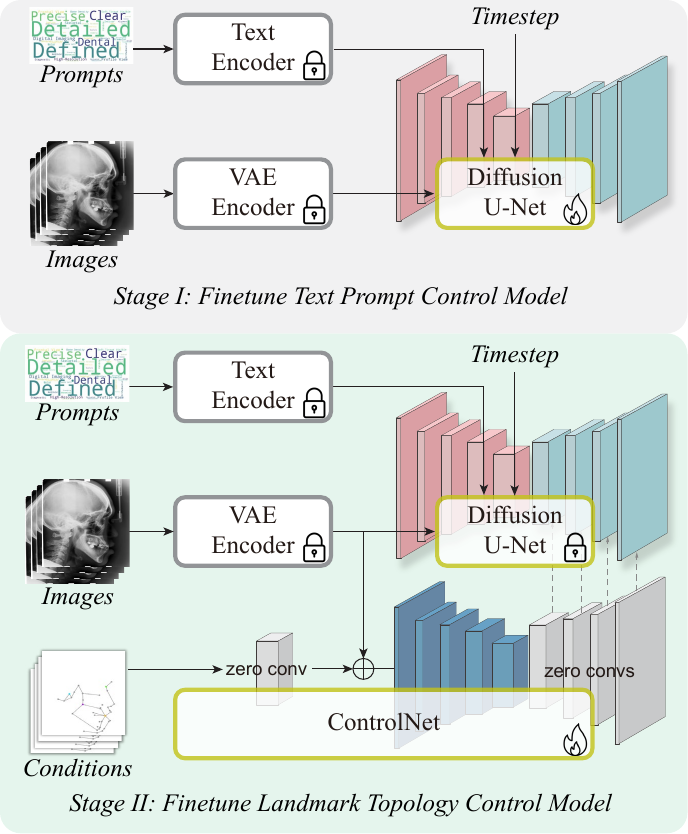}
    \caption{\textbf{The Structure of Cephalometric X-ray Generator and the Adaptation Process.} The adaptation includes two stages: In stage I, we finetune a text-to-image diffusion model to let it adapt to the cephalometric X-ray images domain. In stage II, we freeze the text-to-image diffusion model, connect the ControlNet, load its pre-trained weights, and then finetune it.}
    \label{fig:generation_network}
\end{figure}

Our cephalometric X-ray generation network is constructed of two main components, which are based on the pre-trained Stable Diffusion~\cite{rombach2022high} and ControlNet~\cite{zhang2023adding}. \textbf{1) Text Prompt Control Model:} There is a significant domain gap between natural and medical image-text pairs. To address this, we first finetune the pre-trained Stable Diffusion model using our Prompt-CX dataset. This dataset contains a large and diverse set of medical images paired with text, enabling the model to adapt and optimize the generation of cephalometric X-rays with effective text prompt control.
\textbf{2) Landmark Topology Control Model:} To better align landmark conditions with the generated X-ray images, we introduce a Landmark Topology Control Model based on ControlNet. This model guides the denoising process of the Text Prompt Control Model by capturing topology information between landmarks using zero convolution layers. The output from these layers is then incorporated into the corresponding layers of the Text Prompt Control Model, ensuring more accurate landmark alignment in the generated images.

Figure~\ref{fig:generation_network} shows how we incorporate these two components together for cephalometric X-ray image generation. During the first stage of training, the input image \( I^{b \times c \times h \times w} \) is first encoded into the latent space \( x^{b \times 4 \times 64 \times 64} \) using a frozen, pre-trained VAE encoder. The input text prompt \(p\) is converted into \( k \) tokens and then encoded into embeddings \( h^{b \times k \times 768} \).
\begin{equation}
    x = E_{\text{VAE}}(I),
\end{equation}
\begin{equation}
h = E_{\text{text}}(\mathcal{T}(p)),
\end{equation}
The random Gaussian noise \( \epsilon^{b \times 4 \times 64 \times 64} \) is then added to the latent space \( x \) based on the timestep.
\begin{equation}
    x_{t} = \sqrt{\alpha_t} x_{t-1} + \sqrt{1 - \alpha_t} \epsilon,
\end{equation}
where \(x_{t}\) is the noisy latents at step \(t\), \(\alpha_t\) is a variance schedule.

The noisy latent space, along with the text embeddings and the timestep, are fed into the UNet to predict the noise added at the current timestep. The predicted noise will calculate a loss with the actual noise, which is used to optimize the UNet.
\begin{equation}
    \bar{\epsilon}(x_t, t) = \text{U-Net}(x_{t}, t, h)
\end{equation}
\begin{equation}
   \mathcal{L} = \text{MSE}(\bar{\epsilon}, \epsilon)
\end{equation}
where \(\epsilon_\theta(x_t, t)\) is the predicted noise by the model at step \(t\). 
The finetuned Text Prompt Control Model in Stage I can separately generate cephalometric X-ray images without landmark topology control.

\begin{table*}[ht]
    \renewcommand{\arraystretch}{1.80}
    \centering
    \caption{\textbf{Dataset Overview.} This table details the institution, quantity, resolution, spacing value, and imaging style of the images in the three datasets. * indicates images duplicated with the PKU cephalogram dataset.}
    \begin{tabular}{c|c|c|c|c|c}
        \hline
        \textbf{Dataset} & \textbf{Number} & \textbf{\makecell{Resolution \\ (pixel)}} & \textbf{\makecell{Spacing \\ (mm/pixel)}} & \textbf{Scanners} & \textbf{Samples of different scanners} \\
        \hline
        \hline
        \makecell{PKU \\ Dataset} & 102 & \(1937 \times 2089\) & 0.125 & (a) Planmeca ProMax 3D machine (Finland) & \multirow{3}{*}{\begin{minipage}{0.3\textwidth}\vspace{0.5ex}\includegraphics[width=\textwidth]{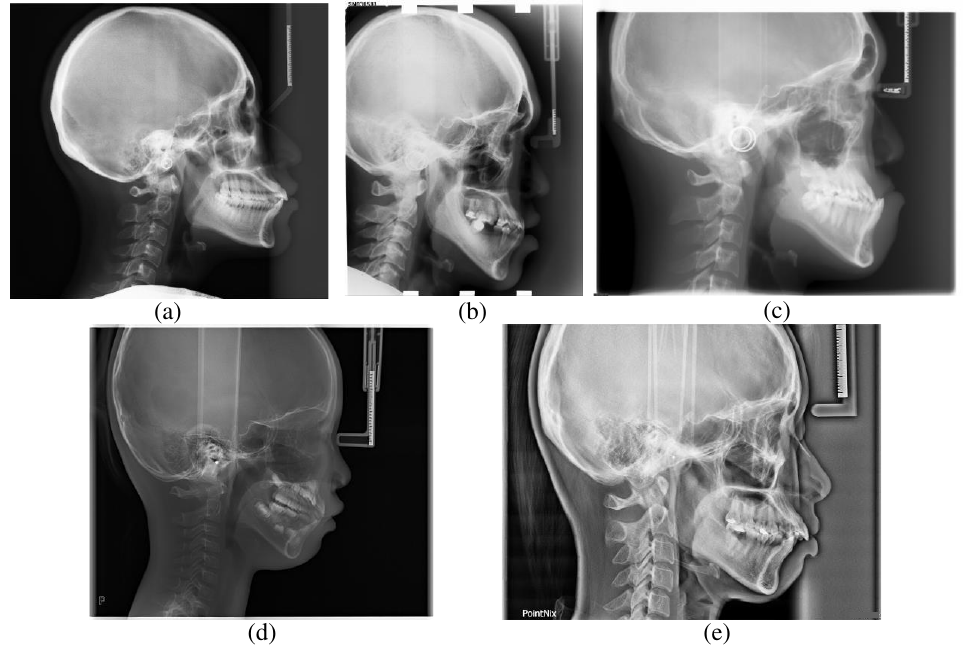} \end{minipage}} \\
        \cline{1-5}
        \multirow{3}{*}{\makecell{CLD \\ Dataset}} & 266 & \(2400 \times 1935\) & 0.1 & \makecell{(b) Soredex CRANEXr Excel Ceph machine \\ (Tuusula, Finland)} & ~ \\
         ~ & 68\textsuperscript{*} & \(1937 \times 2089\) & 0.125 & (a) Planmeca ProMax 3D machine (Finland) \\
         ~ & 66 & \(2304 \times 2880\) & 0.096 & \makecell{(c) Soredex Cranex D Ceph machine \\ (Tuusula, Finland)} & ~ \\
        \cline{1-5}
        \multirow{2}{*}{\makecell{Newly Collected \\ Data}} & 100 & \(2687 \times 2304\) & 0.087 & (d) KAVO X-TREND \\
        ~ & 58 & \(2696 \times 2100\) & 0.089 & (e) PointNix Point800S HD 3D Plus & ~ \\
        \hline
    \end{tabular}
    \label{tab:dataset_overview}
\end{table*}

The Landmark Topology Control Model is more like a plugin of the Text Prompt Control Model. During training, the parameters of the Text Prompt Control Model are fixed, and the Cephalometric Anatomy-Informed Topology is introduced as control conditions. The conditions are passed through a zero convolution and then added to the input noisy latent \( x \), which is subsequently fed into the ControlNet. The structure of ControlNet is largely identical to the UNet module in Stable Diffusion, except that the upsampling layers are replaced with zero convolution layers. The outputs of these zero convolution layers are added to the corresponding layers of the UNet in Stable Diffusion, thereby achieving the desired control effect.
The denoising process of our generator can be represented as:
\begin{equation}
\epsilon_\theta(x_{t-1}, t) = \textit{\textbf{f}}(x_{t}; W_{\text{t2i}}) + \textit{\textbf{z}}(\textit{\textbf{g}}(x_{t} + \textit{\textbf{z}}(c; W_{\text{z1}}); W_{\text{ctrl}}); W_{\text{z2}}),
\end{equation}
where \(\textit{\textbf{f}}(\cdot;W_{\text{t2i}})\) is the Text Prompt Control Model, with the parameters \(W_\text{t2i}\).
\(\textit{\textbf{g}}(\cdot;W_{\text{ctrl}})\) is the Landmark Topology Control Model, with the parameters \(W_\text{ctrl}\).
\(\textit{\textbf{z}}(\cdot; \cdot)\) is the zero convolution layers, 
\(W_\text{z1}\) and \(W_\text{z2}\) are the parameters of the zero convolution layers, \(c\) is the landmark conditions.

\subsection{Cephalometric Landmark Detection Method}
\label{sec:cephalometric_landmark_detection_method}
Thanks to the scale and diversity of the generated data, we can confidently select more powerful detectors without the concern that large vision models will overfit to smaller datasets. To demonstrate the versatility of the generated data across different scaled-detection models, our detection pipeline incorporates various backbone settings, ranging from lightweight to large-scale detection model settings.

\begin{figure}
    \centering
    \includegraphics[scale=0.50]{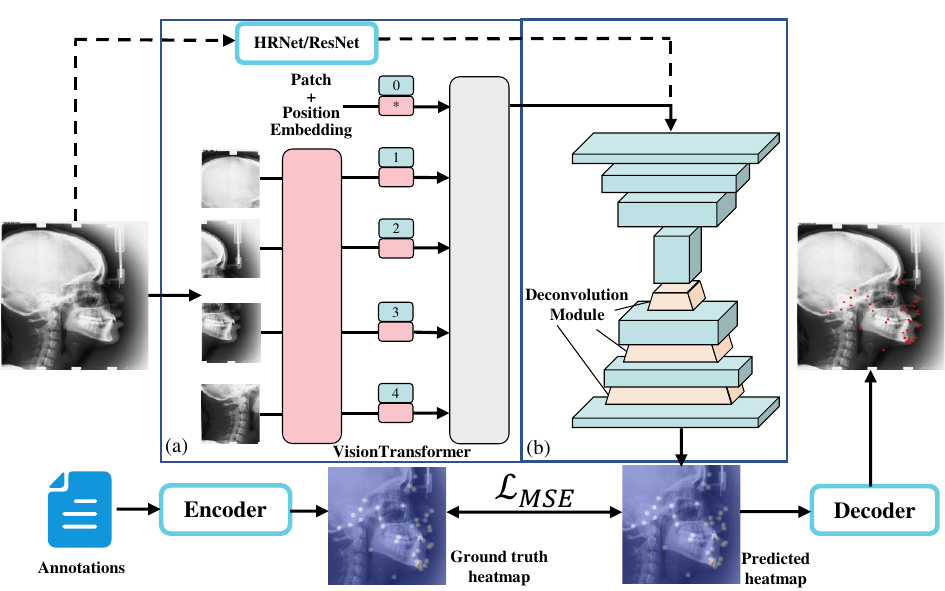}
    \caption{\textbf{The Proposed Cephalometric Landmark Detection Pipeline.} Part (a) is the backbone, which can be replaced by different networks. Part (b) is the head module.}
    \label{fig:detection_pipeline}
\end{figure}

Given a dataset \(D\) containing several cephalometric X-ray images \(I\), each image is associated with ground truth annotations represented as \(L_{gt}\). 
Our pipeline is to train a neural network \(N(I;\theta)\), where the parameters are represented by \(\theta\), and this network provides the predicted landmark coordinates \( L_{pred} \). We adopt the backbone-head framework to construct our detection pipeline, as shown in Figure~\ref{fig:detection_pipeline}.
The encoder will convert each landmark into individual ground truth heatmaps as a supervised objective. The backbone will extract the features of the input image, and then the head module will convert the feature maps into predicted heatmaps and calculate the MSE loss with the ground truth heatmaps. 
The decoder will convert the predicted heatmaps into coordinates. We use \(B\) and \(\mathcal{F}\) to represent the backbone and the feature map extracted through the backbone, respectively, then the neural network \(N(I;\theta)\) mentioned above can be expanded as follows:
\begin{equation}
    N(I;\theta) \Rightarrow Head(\mathcal{F};\theta_2) \Rightarrow Head(B(I;\theta_1);\theta_2).
\end{equation}

The \(\theta_1\) and \(\theta_2\) are the parameters of Backbone and Head. Then, we use \(M_{pred}\) and \(M_{gt}\) to represent the predicted and ground truth heatmaps. The predicted heatmap is the neural network's output, and the ground truth heatmap is obtained through an encoder, which is without trainable parameters. During the training, we actually find the parameters \(\theta\) that minimize the loss between \(M_{pred}\) and \(M_{gt}\)
\begin{equation}
    M_{pred} = N(I;\theta), M_{gt} = \text{Encoder}(L_{gt}),
\end{equation}
\begin{equation}
    \theta^* = \arg\min_\theta \sum_{(I, L_{gt}) \in D} \mathcal{L}_{\text{MSE}}(M_{pred}, M_{gt}) .
\end{equation}

We employ different scales of ResNet~\cite{he2016deep}, HRNet~\cite{wang2020deep}, and VisionTransformer~\cite{dosovitskiy2020image} as our backbones to compare their ability to use generated images. 
Moreover, we use a top-down heatmap head introduced in~\cite{xiao2018simple}, which contains some deconvolutional layers followed by a convolutional layer to generate heatmaps from low-resolution feature maps.

To demonstrate that the generated cephalometric X-ray images can alleviate data scarcity, we generate a large amount of data based on different prompts and landmark controls. 
These data are used to pre-train landmark detection pipelines with various backbone settings. We find that regardless of the parameter size of the backbone network, the pre-trained models show improvements compared to the models without pre-train. 
The results are shown in Section~\ref{sec:experiment}, which proves that the generated data can effectively enhance downstream landmark detection tasks and better transfer the prior knowledge of the Stable Diffusion model to the detection pipeline.

\begin{figure*}
    \centering
    \includegraphics[scale=0.45]{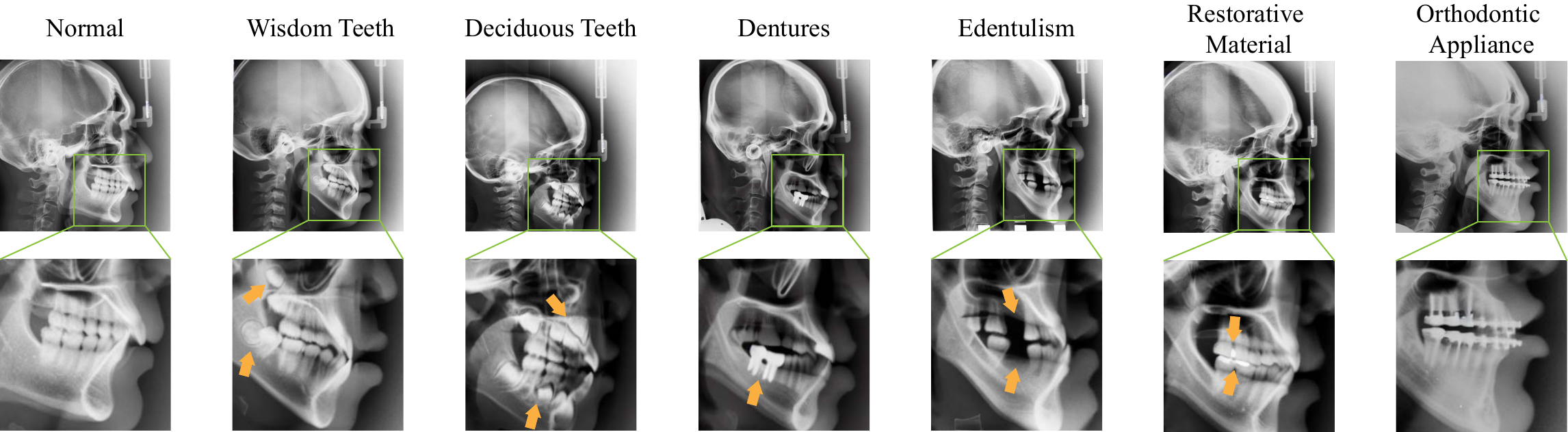}
    \caption{\textbf{The Qualitative Results of the Generated Images with Different Prompts.} We use seven different prompts to generate the images.}
    \label{fig:generated_images}
\end{figure*}

\section{The Proposed Prompt-Cephalometric X-ray (Prompt-CX) Dataset}
\label{sec:datasets}
To train a generative model that is precisely controlled by medical text descriptions and landmark annotations, we introduce the Prompt-Cephalometric X-ray (Prompt-CX) Dataset. This dataset consists of a diverse collection of real cephalometric X-ray images, each accompanied by medical descriptions manually annotated by experienced clinical doctors. Additionally, we provide meticulously labeled landmarks for each image. Table~\ref{tab:dataset_overview} presents an overview of the Prompt-CX dataset.

The Prompt-CX dataset is derived from three distinct data sources. The first is the PKU cephalogram dataset\cite{zeng2021cascaded}, which contains cephalometric images from 102 patients in age from 9 to 53. The second source is the CL-Detection2023 dataset\cite{cld-dataset}, proposed for the MICCAI 2023 Cephalometric Landmark Detection Challenge, comprising 400 training X-ray images. Both of these datasets are representative, encompassing X-ray images from various scanners and distributions. However, they alone are insufficient for training a generator capable of producing realistic X-ray images with diverse features.
To address this limitation, we also introduce newly collected data, which includes a broader variety of features, such as images of children, elderly individuals, and patients with various medical conditions. Specifically, this new dataset consists of 158 fresh lateral cephalometric X-ray images captured using two different scanners. We manually annotated 38 landmarks, ensuring that they are compatible with nearly all clinical cephalometric analysis and are aligned with the other datasets.

To enhance the applicability of these datasets for our generative work, we introduce new annotations by adding medical text descriptions to 592 images. Specifically, we enrich the semantic information of each image with 50 distinct descriptions. Moreover, for X-ray images captured by different scanners with varying imaging styles, we include the scanner names as additional semantic information in the annotations. 

\begin{figure}[t]
    \centering
    \includegraphics[scale=0.2]{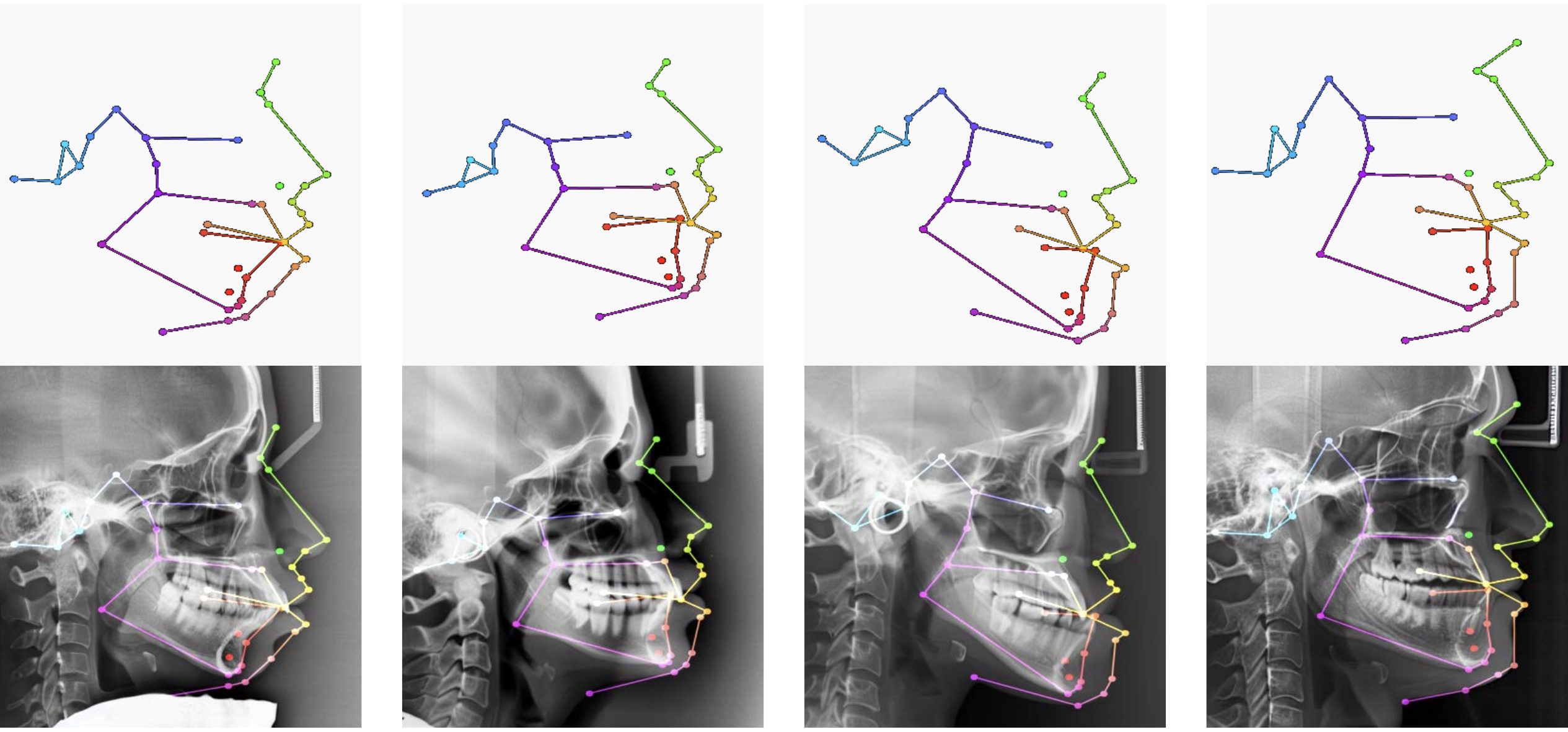}
    \caption{\textbf{The Qualitative Results of the Generated Images with Different Landmark Conditions.} We illustrate four landmark conditions, generating cephalometric X-ray images captured by four scanners based on these landmarks.}
    \label{fig:controlnet_vis}
\end{figure}

\section{Experimental Results}
\label{sec:experiment}
\subsection{Cephalometric X-ray Images Generation}
\renewcommand{\arraystretch}{1.5}
\begin{table*}[ht]
    \centering
    \caption{\textbf{Detection performance of three phases on the whole dataset.}}
    \setlength{\tabcolsep}{13pt}
    \resizebox{\linewidth}{!}{
    \begin{tabular}{c|c|c|c|c|c|c|c|c}
        \toprule
        \multirow{2}{*}{\textbf{Backbones}} & \multirow{2}{*}{\textbf{\makecell{Params \\ (M)}}} & \multirow{2}{*}{\textbf{\makecell{FLOPs \\ (G)}}} & \multicolumn{2}{c|}{\textbf{Without Pre-train}} & \multicolumn{2}{c|}{\textbf{Pre-train Only}} & \multicolumn{2}{c}{\textbf{Pre-train + Finetune}} \\
        \cline{4-9}
        ~ & ~ & ~ & \textbf{\makecell{MRE ± SD $\downarrow$ \\ (mm) }} &  \textbf{\makecell{SDR 2mm $\uparrow$ \\ (\%) }} &  \textbf{\makecell{MRE ± SD $\downarrow$ \\ (mm) }} &  \textbf{\makecell{SDR 2mm $\uparrow$ \\ (\%) }} &  \textbf{\makecell{MRE ± SD $\downarrow$ \\ (mm) }} & \textbf{\makecell{SDR 2mm $\uparrow$ \\ (\%) }} \\
        \hline
        \hline
        ResNet-50 & 34.0 & 5.5 & 1.753 ± 1.807 & 71.147 & 1.941 ± 1.972 & 66.541 & 1.673 ± 1.895 & 74.624(+3.117)\\
        ResNet-101 & 53.0 & 9.1 & 1.738 ± 1.898 & 72.368 & 1.959 ± 1.979 & 65.038 & 1.639 ± 1.818 &	75.157(+2.789)\\
        ResNet-152 & 68.6 & 12.8 & 1.728 ± 1.833 & 72.525 & 1.925 ± 1.969 & 66.792 & 1.632 ± 1.824 & 75.940(+3.415)\\
        \hline
        HRNet-w32 & 28.5 & 7.7 & 2.017 ± 1.958 & 63.878 & 2.024 ± 2.009 & 62.845 & 1.911 ± 1.939 & 66.197(+2.319)\\
        HRNet-w48 & 63.6 & 15.7 & 1.684 ± 1.794 & 74.029 & 1.949 ± 2.188 & 66.886 & 1.668 ± 1.898 & 75.940(+1.911)\\
        \hline
        ViT-base & 90.6 & 17.9 & 1.592 ± 1.926 & 75.345 & 1.748 ± 1.875 & 70.959 & 1.464 ± 1.784 & 78.853(+3.508)\\
        ViT-large & 309.0 & 59.8 & 1.589 ± 1.883 & 75.470 & 1.633 ± 1.805 & 73.308 & 1.452 ± 1.881 & 79.010(+3.540)\\
        ViT-huge & 638.0 & 123.0 & \textbf{1.550 ± 1.849} & \textbf{75.752} & \textbf{1.614 ± 1.793} & \textbf{73.841} & \textbf{1.365 ± 1.710} & \textbf{82.206(+6.454)}\\
        \bottomrule
    \end{tabular}}
    \label{tab:detection_result_1}
\end{table*}

\begin{figure}
    \centering
    \includegraphics[scale=0.2]{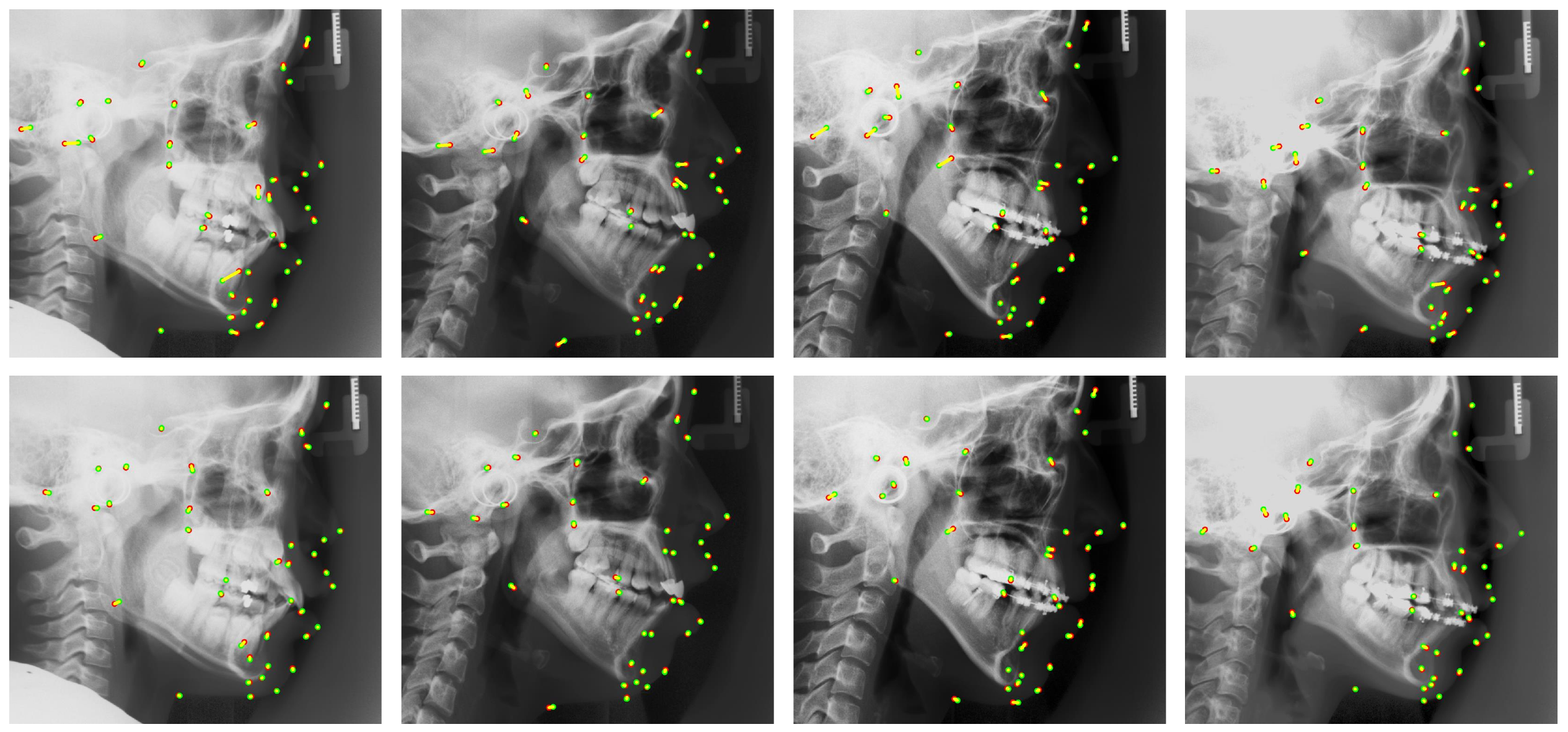}
    \caption{\textbf{The Qualitative Detection Results in the Test Set.} Green dots represent ground truth landmarks, and red dots show predicted landmarks, with yellow lines connecting them. The first row shows results without pre-training, and the second row shows results with pre-training and fine-tuning. From left to right, the features are deciduous teeth, dentures, missing teeth, and orthodontic appliances.}
    \label{fig:corner_cases_vis}
\end{figure}

\noindent\textbf{Implementation Detail:} The training of the generation network includes two stages; both are implemented based on the Diffusers~\cite{von-platen-etal-2022-diffusers}. 
The Diffusion U-Net is composed of 12 blocks for encoding, a middle block, and 12 blocks for decoding. 
During training, we use 476 real images from the Prompt-CX dataset, resized to \(512 \times 512\) with cropping for data augmentation. First, we finetune the Text Prompt Control Model, initializing the generation pipeline with pre-trained weights from Hugging Face~\cite{rombach2022high}. Training uses an NVIDIA RTX 4090 with a learning rate of \(1 \times 10^{-5}\), with an effective batch size of 4. This stage lasts 20,000 steps. Then, we freeze the Text Prompt Control Model and train the Landmark Topology Control Model with designed landmarks as conditions, using pre-trained ControlNet weights~\cite{zhang2023adding}. 
This stage is with a batch size of 64 and lasts 15,000 steps.

\noindent\textbf{Results:} Figure~\ref{fig:generated_images} shows images by different PDG prompts, demonstrating accurate image generation on prompts. Figure~\ref{fig:controlnet_vis} displays results with different landmarks as controls. The PDG-generated prompt is: \textit{Cephalometric X-ray, Permanent teeth, Real teeth, Dentate, Without dental fillings, Without braces,} representing X-ray images with no major oral abnormalities. Each image is generated with 150 inference steps.

\renewcommand{\arraystretch}{1.5}

\begin{table*}[ht]
    \centering
    \caption{\textbf{Detection performance of the minority features in the test set.}}
    \resizebox{\textwidth}{!}{
    \setlength{\tabcolsep}{10pt}
    \begin{tabular}{c|c|c|c|c|c|c|c|c|c|c}
        \toprule
        \multirow{3}{*}{\textbf{Features}} & \multicolumn{5}{c|}{\textbf{Without Pre-train}} & \multicolumn{5}{c}{\textbf{Pre-train + Finetune}} \\
        \cline{2-11}
        ~ & \multirow{2}{*}{\textbf{\makecell{MRE ± SD $\downarrow$ \\ (mm)}}} & \multicolumn{4}{c|}{\textbf{SDR (\%) $\uparrow$}} & \multirow{2}{*}{\textbf{\makecell{MRE ± SD $\downarrow$ \\ (mm)}}} & \multicolumn{4}{c}{\textbf{SDR (\%) $\uparrow$}} \\
        \cline{3-6}
        \cline{8-11}
        ~ & ~ & \textbf{2mm} & \textbf{2.5mm} & \textbf{3mm} & \textbf{4mm} & ~ & \textbf{2mm} & \textbf{2.5mm} & \textbf{3mm} & \textbf{4mm} \\
        \hline
        \hline
        Deciduous teeth & 1.596 ± 2.520	& 75.376 & 85.902 &	89.850 & 94.361 & 1.415 ± 2.407 & 82.707(+7.331) & 88.534 & 91.541 & 95.865 \\
        Dentures & 1.728 ± 1.775 & 71.330 & 79.501 & 83.518 & 91.413 & 1.507 ± 1.934 & 78.809(+7.479) & 84.903 & 89.335 & 93.629 \\
        Orthodontic appliance & 1.788 ± 1.676 &	69.549 & 78.195 & 82.331 & 91.729 &	1.441 ± 1.279 &	78.571(+9.022) & 84.586 & 88.346 & 94.361 \\
        Missing teeth & 1.795 ± 1.650 & 69.070 & 77.630 & 82.230 & 91.440 & 1.566 ± 1.490 & 76.310(+7.240) & 83.220 & 87.170 & 94.408 \\
        \bottomrule
    \end{tabular}}
    \label{tab:corner_case_result}
\end{table*}

\subsection{Cephalometric Landmark Detection}
\noindent\textbf{Implementation Detail:}
Regarding the cephalometric landmark detection pipeline, we conduct three main experiments. All the 592 images are split into three parts: training dataset (476 images), validation dataset (32 images), and test dataset (84 images). 
The pipeline is implemented based on the MMPose library. All experiments are conducted on an NVIDIA RTX 4090 using the Adam optimizer with a learning rate of $5 \times 10^{-4}$ and trained for 250 epochs, with validation performed once per epoch to select the best-performing model. The batch size is set to 5, and the input image size is \(512 \times 512\). 

\noindent\textbf{Results:} From Table~\ref{tab:detection_result_1} shows that among the eight backbones compared, ViT-huge achieves the best performance in all experiments. The finetuned ViT-huge attains an overall MRE of $1.365 \pm 1.710$~mm and an SDR (within 2~mm) of 82.206\%. Compared to the HRNet-w32 baseline (without pre-training), the finetuned ViT-huge reduces the MRE by 0.652~mm, the standard deviation by 0.248~mm, and increases the SDR by 18.328\%. Although all backbones benefit from pre-training and fine-tuning, the improvement is most significant for ViT-huge, which has the highest parameter count and complexity. Compared to ViT-huge without pre-training, its MRE decreases by 0.185~mm, standard deviation by 0.139~mm, and SDR increases by 6.454\%, demonstrating that using generated images not only improves detection accuracy but also minimizes prediction error fluctuations across landmarks.

\subsection{Ablation Study}
Table~\ref{tab:abl_1} presents the results of ablation experiments on different training strategies. Initially, the model is trained only with synthetic data in quantities equivalent to, four times and eight times the size of the real training dataset. Using a ViT-huge backbone and the same training setup, the model achieves notable performance even with limited synthetic data. Specifically, the SDR within 2mm metric reaches nearly 70\%, comparable to the performance of training ResNet-50 with real data. As the amount of synthetic data increases, the performance improves further, with the SDR within 2mm metric reaching 73.841\%. Next, the experiments are conducted by mixing synthetic and real data in varying proportions. The results demonstrate that incorporating a substantial amount of synthetic data significantly enhances the model’s performance. Compared to training solely with real data, the SDR within 2mm metric improved by nearly 10\%. These findings highlight the effectiveness of synthetic data in improving robustness.

Table~\ref{tab:abl_2} summarizes ablation experiments on different modules. These experiments use 3,808 synthetic images generated with various module combinations and the detection pipeline. The baseline model relies on discrete landmarks, without topological connections, color variations, or augmentations, and uses a single general text prompt for all images. Adding the MIRA and AIT modules independently improves the SDR by 11.230\% and 20.012\%, respectively, showing the importance of augmentations and topology in enhancing image quality. Introducing the PDG module further boosts performance by generating more diverse images, achieving an SDR of 68\%. Combining all three modules achieved the best results, with an MRE of $1.614 \pm 1.793$ and an SDR within 2mm of 73.841\%.

\begin{table}[t]
    \centering
    \caption{{\textbf{The Ablative Study of Different Training Strategy.}}}
    \setlength{\tabcolsep}{15pt}
    \resizebox{0.98\linewidth}{!}{
    \begin{tabular}{c|c|c|c}
    \toprule
    \textbf{Real} & \textbf{Syn} & \textbf{\makecell{MRE ± SD $\downarrow$ \\ (mm)}} & \textbf{\makecell{SDR 2mm $\uparrow$ \\ (\%) }} \\
    \hline
    \hline
    0 & 476 & 1.791 ± 1.823 & 69.986 \\
    0 & 1904 & 1.698 ± 1.716 & 72.935 \\
    0 & 3808 & 1.614 ± 1.793 & 73.841 \\
    476 & 0 & 1.550 ± 1.849 & 75.752 \\
    476 & 476 & 1.162 ± 1.605 & 84.807 \\
    476 & 1904 & 1.122 ± 1.535 & 85.113 \\
    \bottomrule
    \end{tabular}}
    \label{tab:abl_1}
\end{table}

\begin{table}[t]
    \centering
    \caption{\textbf{Expert Study Results}.}
    \resizebox{0.46\textwidth}{!}{
    \begin{tabular}{c|c|c|c|c}
        \toprule
        \multirow{2}{*}{\textbf{Data}} & \multicolumn{2}{c|}{\textbf{Correspondence}} & \multicolumn{2}{c}{\textbf{Quality}} \\ 
        \cline{2-5}
        ~ & \textbf{Average Points} & \textbf{$\geq$ 4 Points} & \textbf{Average Points} & \textbf{$\geq$ 4 Points} \\
        \hline
        \hline
        Real & 4.92 & 100\% & 4.96& 100\% \\
        Sync & 4.74 & 92\% & 4.59  & 98\% \\
        \bottomrule
    \end{tabular}
    }
    \label{tab:human_evaluation}
\end{table}

\begin{table}[t]
    \centering
    \caption{\textbf{The Ablative Study of Different Modules.}}
    \setlength{\tabcolsep}{15pt}
    \resizebox{0.98\linewidth}{!}{
    \begin{tabular}{c|c|c|c|c}
    \toprule
    \textbf{MIRA} & \textbf{AIT} & \textbf{PDG} & \textbf{\makecell{MRE ± SD $\downarrow$ \\ (mm)}} & \textbf{\makecell{SDR 2mm $\uparrow$ \\ (\%) }} \\
    \hline
    \hline
    ~ & ~ & ~ & 3.662 ± 4.413 & 45.288 \\
    \checkmark & ~ & ~ & 2.309 ± 2.103 & 56.518 \\
    ~ & \checkmark & ~ & 2.119 ± 1.968 & 65.300 \\
    \checkmark & \checkmark & ~ & 2.111 ± 2.030 & 61.460 \\
    \checkmark & ~ & \checkmark & 1.799 ± 1.729 & 68.849 \\
    \checkmark & \checkmark & \checkmark & 1.614 ± 1.793 & 73.841 \\
    \bottomrule
    \end{tabular}}
    \label{tab:abl_2}
\end{table}

\subsection{Expert Study}
We organized an expert study to evaluate the clinical accuracy of generated images and how well augmentations represent real pathological variations. A total of 100 generated and 100 real images were randomly selected, resized to \(512 \times 512\), shuffled, and uniformly numbered. For real images, we manually annotated landmarks and text descriptions. For synthetic images, the generation conditions (with topology information removed) and text prompts were annotated on the images for expert evaluation. Each image was assessed by two doctors. The evaluation focused on two aspects: the correspondence of the landmarks and text descriptions, and whether the image quality met clinical standards. Both aspects were rated on a scale of 1 to 5, from poor to excellent. Table~\ref{tab:human_evaluation} shows the evaluation results. We observe that over 90\% of the synthetic images received scores above 4, indicating high quality with the conditions and clinical-grade image quality.

\subsection{Discussion on the Generalization Ability}
Datasets typically reflect the characteristics of specific populations over time, making it challenging to quickly accumulate large amounts of data for minority groups. For example, in our Prompt-CX dataset, among 592 X-rays from three different sources, only 68 images depict children (i.e., deciduous teeth), 42 images show missing teeth, and 30 images include orthodontic appliances, resulting in insufficient training data for these cases. However, generating synthetic images can cost-effectively augment these scarce clinical scenarios. We generate a large number of images for the underrepresented groups, achieving breakthroughs in landmark recognition performance. For instance, for the four features with significantly rare distributions—``Deciduous teeth'', ``Dentures'', ``Orthodontic appliances'', and ``Missing teeth''—we construct subsets from the test dataset and evaluate performance using ViT-huge as the backbone. 
In Table~\ref{tab:corner_case_result}, the performance improves significantly with the use of generated images, and some qualitative results are shown in Figure~\ref{fig:corner_cases_vis}. For these features, improvements of over 7\% are observed for ViT-huge pre-trained with synthetic data. Thus, using generative images for pre-training greatly enhances the generalization ability of the detection model.

\vspace{2mm}

\section{Conclusion}
\label{sec:conclusion}
In this study, we address the challenges of data scarcity and labor-intensive annotation in cephalometric X-ray landmark detection models by proposing a method to generate synthetic cephalometric X-ray images for pre-training, enhancing landmark detection performance. We collect and annotate datasets with experienced medical experts and supplement them with textual semantic descriptions. Based on anatomical relationships, we design topological connections for the landmarks to train the generative model. We design a two-stage finetuning method to optimize the use of limited samples. Using anatomical rules, we create various landmarks and textual prompts to generate a rich mounting of diffusion-generated images. We then evaluate the performance of pre-training on eight different types and scales of backbone models. The results demonstrate that pre-training with synthetic images improves both the Success Detection Rate and Mean Radial Error. Notably, larger models exhibit significant improvements in landmark detection tasks, effectively leveraging generated images to learn richer prior knowledge.

\vspace{5mm}


\begin{thebibliography}{00}
\bibitem{kielczykowski2023application}
M.~Kie{\l}czykowski, K.~Kami{\'n}ski, K.~Perkowski, M.~Zadurska, and E.~Czochrowska, ``Application of artificial intelligence (ai) in a cephalometric analysis: A narrative review,'' \emph{Diagnostics}, vol.~13, no.~16, p. 2640, 2023.

\bibitem{03}
A.~Sharma and G.~Hamarneh, ``Missing mri pulse sequence synthesis using multi-modal generative adversarial network,'' \emph{IEEE Trans. Med. Imag.}, vol.~39, no.~4, pp. 1170--1183, 2019.

\bibitem{rauniyar2023artificial}
S.~Rauniyar, S.~Jena, N.~Sahoo, P.~Mohanty, and B.~P. Dash, ``Artificial intelligence and machine learning for automated cephalometric landmark identification: A meta-analysis previewed by a systematic review,'' \emph{Cureus}, vol.~15, no.~6, 2023.

\bibitem{04}
T.~Zhou, H.~Fu, G.~Chen, J.~Shen, and L.~Shao, ``Hi-net: hybrid-fusion network for multi-modal mr image synthesis,'' \emph{IEEE Trans. Med. Imag.}, vol.~39, no.~9, pp. 2772--2781, 2020.

\bibitem{05}
Y.~Huang, L.~Shao, and A.~F. Frangi, ``Cross-modality image synthesis via weakly coupled and geometry co-regularized joint dictionary learning,'' \emph{IEEE Trans. Med. Imag.}, vol.~37, no.~3, pp. 815--827, 2017.

\bibitem{wang2015evaluation}
C.-W. Wang, C.-T. Huang, M.-C. Hsieh, C.-H. Li, S.-W. Chang, W.-C. Li, R.~Vandaele, R.~Mar{\'e}e, S.~Jodogne, P.~Geurts \emph{et~al.}, ``Evaluation and comparison of anatomical landmark detection methods for cephalometric x-ray images: a grand challenge,'' \emph{IEEE Trans. Med. Imag.}, vol.~34, no.~9, pp. 1890--1900, 2015.

\bibitem{madani2018chest}
A.~Madani, M.~Moradi, A.~Karargyris, and T.~Syeda-Mahmood, ``Chest x-ray generation and data augmentation for cardiovascular abnormality classification,'' in \emph{Proc. SPIE}, vol. 10574, SPIE, 2018, pp. 415--420.

\bibitem{wang2016benchmark}
C.-W. Wang, C.-T. Huang, J.-H. Lee, C.-H. Li, S.-W. Chang, M.-J. Siao, T.-M. Lai, B.~Ibragimov, T.~Vrtovec, O.~Ronneberger \emph{et~al.}, ``A benchmark for comparison of dental radiography analysis algorithms,'' \emph{Med. Image Anal.}, vol.~31, pp. 63--76, 2016.

\bibitem{karbhari2021generation}
Y.~Karbhari, A.~Basu, Z.~W. Geem, G.-T. Han, and R.~Sarkar, ``Generation of synthetic chest x-ray images and detection of covid-19: A deep learning based approach,'' \emph{Diagnostics}, vol.~11, no.~5, p. 895, 2021.

\bibitem{huang2024diverse}
K.~Huang, X.~Ma, Z.~Zhang, Y.~Zhang, S.~Yuan, H.~Fu, and Q.~Chen, ``Diverse data generation for retinal layer segmentation with potential structure modelling,'' \emph{IEEE Trans. Med. Imag}, 2024.

\bibitem{goodfellow2014generative}
I.~Goodfellow, J.~Pouget-Abadie, M.~Mirza, B.~Xu, D.~Warde-Farley, S.~Ozair, A.~Courville, and Y.~Bengio, ``Generative adversarial nets,'' \emph{Proc. Adv. Neural Inf. Process. Syst.}, vol.~27, 2014.

\bibitem{kwon2019generation}
G.~Kwon, C.~Han, and D.-s. Kim, ``Generation of 3d brain mri using auto-encoding generative adversarial networks,'' in \emph{Proc. Int. Conf. Med. Image Comput. Comput.-Assist. Intervent.}\hskip 1em plus 0.5em minus 0.4em\relax Springer, 2019, pp. 118--126.

\bibitem{skandarania2021gans}
Y.~Skandarania, P.-M. Jodoinc, and A.~Lalandea, ``Gans for medical image synthesis: An empirical study,'' \emph{arXiv:2105.05318}, 2021.

\bibitem{rombach2022high}
R.~Rombach, A.~Blattmann, D.~Lorenz, P.~Esser, and B.~Ommer, ``High-resolution image synthesis with latent diffusion models,'' in \emph{Proc. IEEE Conf. Comput. Vis. Pattern Recognit. (CVPR)}, 2022, pp. 10\,684--10\,695.

\bibitem{wang2020deep}
J.~Wang, K.~Sun, T.~Cheng, B.~Jiang, C.~Deng, Y.~Zhao, D.~Liu, Y.~Mu, M.~Tan, X.~Wang \emph{et~al.}, ``Deep high-resolution representation learning for visual recognition,'' \emph{IEEE Trans. Pattern Anal. Mach. Intell.}, vol.~43, no.~10, pp. 3349--3364, 2020.

\bibitem{weber2023cascaded}
T.~Weber, M.~Ingrisch, B.~Bischl, and D.~R{\"u}gamer, ``Cascaded latent diffusion models for high-resolution chest x-ray synthesis,'' in \emph{Proc. PAKDD}.\hskip 1em plus 0.5em minus 0.4em\relax Springer, 2023, pp. 180--191.

\bibitem{shentu2024cxr}
J.~Shentu and N.~Al~Moubayed, ``Cxr-irgen: An integrated vision and language model for the generation of clinically accurate chest x-ray image-report pairs,'' in \emph{Proc. IEEE Winter Conf. Appl. Comput. Vis}, 2024, pp. 5212--5221.

\bibitem{liang2023pie}
K.~Liang, X.~Cao, K.-D. Liao, T.~Gao, Z.~Chen, and T.~Nama, ``Pie: Simulating disease progression via progressive image editing,'' \emph{arXiv:2309.11745}, 2023.


\bibitem{zhan2024medm2g}
C.~Zhan, Y.~Lin, G.~Wang, H.~Wang, and J.~Wu, ``Medm2g: Unifying medical multi-modal generation via cross-guided diffusion with visual invariant,'' in \emph{Proc. IEEE Conf. Comput. Vis. Pattern Recognit. (CVPR)}, 2024, pp. 11\,502--11\,512.

\bibitem{zhang2023adding}
L.~Zhang, A.~Rao, and M.~Agrawala, ``Adding conditional control to text-to-image diffusion models,'' in \emph{Proc. IEEE Conf. Comput. Vis. Pattern Recognit. (CVPR)}, 2023, pp. 3836--3847.

\bibitem{kaleta2023lc}
J.~Kaleta, D.~Dall'alba, S.~Plotka, and P.~Korzeniowski, ``Lc-sd: Realistic endoscopic image generation with limited training data,'' in \emph{Adv. Neural Inf. Process. Syst. Workshop}, 2023.

\bibitem{pinaya2023generative}
W.~H. Pinaya, M.~S. Graham, E.~Kerfoot, P.-D. Tudosiu, J.~Dafflon, V.~Fernandez, P.~Sanchez, J.~Wolleb, P.~F. da~Costa, A.~Patel \emph{et~al.}, ``Generative ai for medical imaging: extending the monai framework,'' \emph{arXiv:2307.15208}, 2023.


\bibitem{levy1986knowledge}
A.~Levy-Mandel, A.~Venetsanopoulos, and J.~Tsotsos, ``Knowledge-based landmarking of cephalograms,'' \emph{Comput. Biomed. Res.}, vol.~19, no.~3, pp. 282--309, 1986.

\bibitem{mosleh2008image}
M.~A. Mosleh, M.~S. Baba, N.~Himazian, and B.~M. AL-Makramani, ``An image processing system for cephalometric analysis and measurements,'' in \emph{Int. Symp. Inf. Technol.}, vol.~4.\hskip 1em plus 0.5em minus 0.4em\relax IEEE, 2008, pp. 1--8.

\bibitem{pouyan2010cephalometric}
A.~A. Pouyan and M.~Farshbaf, ``Cephalometric landmarks localization based on histograms of oriented gradients,'' in \emph{Proc. Int. Conf. Signal Image Process.}\hskip 1em plus 0.5em minus 0.4em\relax IEEE, 2010, pp. 1--6.

\bibitem{ibragimov2014automatic}
B.~Ibragimov, B.~Likar, F.~Pernus, and T.~Vrtovec, ``Automatic cephalometric x-ray landmark detection by applying game theory and random forests,'' in \emph{Proc. IEEE 11th Int. Symp. Biomed. Imag. (ISBI)}, 2014, pp. 1--8.

\bibitem{tim2015fully}
C.~Tim, F.~Cootes \emph{et~al.}, ``Fully automatic cephalometric evaluation using random forest regression-voting,'' in \emph{Proc. IEEE 12th Int. Symp. Biomed. Imag. (ISBI)}, 2015, pp. 16--19.

\bibitem{lindner2016fully}
C.~Lindner, C.-W. Wang, C.-T. Huang, C.-H. Li, S.-W. Chang, and T.~F. Cootes, ``Fully automatic system for accurate localisation and analysis of cephalometric landmarks in lateral cephalograms,'' \emph{Sci. Rep.}, vol.~6, no.~1, p. 33581, 2016.

\bibitem{wu2023revisiting}
Q.~Wu, S.~Y. Yeo, Y.~Chen, and J.~Liu, ``Revisiting cephalometric landmark detection from the view of human pose estimation with lightweight super-resolution head,'' \emph{arXiv:2309.17143}, 2023.

\bibitem{lee2019deep}
C.~Lee, C.~Tanikawa, J.-Y. Lim, and T.~Yamashiro, ``Deep learning based cephalometric landmark identification using landmark-dependent multi-scale patches,'' \emph{arXiv:1906.02961}, 2019.

\bibitem{chen2019cephalometric}
R.~Chen, Y.~Ma, N.~Chen, D.~Lee, and W.~Wang, ``Cephalometric landmark detection by attentive feature pyramid fusion and regression-voting,'' in \emph{Proc. Int. Conf. Med. Image Comput. Comput.-Assist. Intervent.}\hskip 1em plus 0.5em minus 0.4em\relax Springer, 2019, pp. 873--881.

\bibitem{jiang2022cephalformer}
Y.~Jiang, Y.~Li, X.~Wang, Y.~Tao, J.~Lin, and H.~Lin, ``Cephalformer: incorporating global structure constraint into visual features for general cephalometric landmark detection,'' in \emph{Proc. Int. Conf. Med. Image Comput. Comput.-Assist. Intervent.}\hskip 1em plus 0.5em minus 0.4em\relax Springer, 2022, pp. 227--237.

\bibitem{zhu2023uod}
H.~Zhu, Q.~Quan, Q.~Yao, Z.~Liu, and S.~K. Zhou, ``Uod: Universal one-shot detection of anatomical landmarks,'' in \emph{Proc. Int. Conf. Med. Image Comput. Comput.-Assist. Intervent.}\hskip 1em plus 0.5em minus 0.4em\relax Springer, 2023, pp. 24--34.

\bibitem{tweed1946frankfort}
C.~H. Tweed, ``The frankfort-mandibular plane angle in orthodontic diagnosis, classification, treatment planning, and prognosis,'' \emph{Am. J. Orthod. Oral Surgery}, vol.~32, no.~4, pp. 175--230, 1946.

\bibitem{steiner1953cephalometrics}
C.~C. Steiner, ``Cephalometrics for you and me,'' \emph{Am. J. Orthod.}, vol.~39, no.~10, pp. 729--755, 1953.

\bibitem{radford2021learning}
A.~Radford, J.~W. Kim, C.~Hallacy, A.~Ramesh, G.~Goh, S.~Agarwal, G.~Sastry, A.~Askell, P.~Mishkin, J.~Clark \emph{et~al.}, ``Learning transferable visual models from natural language supervision,'' in \emph{Int. Conf. on Mach. Learn.}\hskip 1em plus 0.5em minus 0.4em\relax PMLR, 2021, pp. 8748--8763.

\bibitem{he2016deep}
K.~He, X.~Zhang, S.~Ren, and J.~Sun, ``Deep residual learning for image recognition,'' in \emph{Proc. IEEE Conf. Comput. Vis. Pattern Recognit. (CVPR)}, 2016, pp. 770--778.

\bibitem{xiao2018simple}
B.~Xiao, H.~Wu, and Y.~Wei, ``Simple baselines for human pose estimation and tracking,'' in \emph{Proc. Eur. Conf. Comput. Vis. (ECCV)}, 2018, pp. 466--481.

\bibitem{zeng2021cascaded}
M.~Zeng, Z.~Yan, S.~Liu, Y.~Zhou, and L.~Qiu, ``Cascaded convolutional networks for automatic cephalometric landmark detection,'' \emph{Med. Image Anal.}, vol.~68, p. 101904, 2021.

\bibitem{cld-dataset}
W.~Ching-Wei, H.~Bingsheng, Z.~Hongyuan, M.~Hikam, C.~Jun, D.~Juan, and L.~Xuguang, ``Cephalometric landmark detection in lateral x-ray images 2023,'' \url{https://cl-detection2023.grand-challenge.org/}, 2023.

\bibitem{von-platen-etal-2022-diffusers}
P.~von Platen, S.~Patil, A.~Lozhkov, P.~Cuenca, N.~Lambert, K.~Rasul, M.~Davaadorj, D.~Nair, S.~Paul, W.~Berman, Y.~Xu, S.~Liu, and T.~Wolf, ``Diffusers: State-of-the-art diffusion models,'' \url{https://github.com/huggingface/diffusers}, 2022.

\bibitem{quan2022images}
Q.~Quan, Q.~Yao, J.~Li, and S.~K. Zhou, ``Which images to label for few-shot medical landmark detection?'' in \emph{Proc. IEEE Conf. Comput. Vis. Pattern Recognit. (CVPR)}, 2022, pp. 20\,606--20\,616.

\bibitem{wolf2019huggingface}
T.~Wolf, L.~Debut, V.~Sanh, J.~Chaumond, C.~Delangue, A.~Moi, P.~Cistac, T.~Rault, R.~Louf, M.~Funtowicz \emph{et~al.}, ``Huggingface's transformers: State-of-the-art natural language processing,'' \emph{arXiv:1910.03771}, 2019.

\bibitem{dosovitskiy2020image}
A.~Dosovitskiy, L.~Beyer, A.~Kolesnikov, D.~Weissenborn, X.~Zhai, T.~Unterthiner, M.~Dehghani, M.~Minderer, G.~Heigold, S.~Gelly \emph{et~al.}, ``An image is worth 16x16 words: Transformers for image recognition at scale,'' \emph{arXiv:2010.11929}, 2020.


\bibitem{paddenberg2021floating}
E.~Paddenberg, P.~Proff, and C.~Kirschneck, ``Floating norms for individualising the anb angle and the wits appraisal in orthodontic cephalometric analysis based on guiding variables,'' \emph{Journal of Orofacial Orthopedics}, vol.~84, no.~1, p.~10, 2021.

\bibitem{chen2021synthetic}
R.~J. Chen, M.~Y. Lu, T.~Y. Chen, D.~F. Williamson, and F.~Mahmood, ``Synthetic data in machine learning for medicine and healthcare,'' \emph{Nature Biomedical Engineering}, vol.~5, no.~6, pp. 493--497, 2021.

\bibitem{garcea2023data}
F.~Garcea, A.~Serra, F.~Lamberti, and L.~Morra, ``Data augmentation for medical imaging: A systematic literature review,'' \emph{Computers in Biology and Medicine}, vol. 152, p. 106391, 2023.

\bibitem{pezoulas2024synthetic}
V.~C. Pezoulas, D.~I. Zaridis, E.~Mylona, C.~Androutsos, K.~Apostolidis, N.~S. Tachos, and D.~I. Fotiadis, ``Synthetic data generation methods in healthcare: A review on open-source tools and methods,'' \emph{Computational and Structural Biotechnology Journal}, 2024.

\bibitem{pengfeiguo2024addressing}
G.~Pengfei, Y.~Dong, Z.~Can, and X.~Daguang, ``Addressing medical imaging limitations with synthetic data generation,'' \emph{NVidia Tech. Blog}, 2024.

\end{thebibliography}
\end{document}